\documentclass[11pt,a4paper]{article}
\usepackage[hyperref]{emnlp2020}
\usepackage{times}
\usepackage{latexsym}
\usepackage{mylingmacros}
\usepackage{graphicx}

\newcommand{\conn[1]}{\underline{#1}}
\newcommand{\argi[1]}{\textit{#1}}
\newcommand{\argii[1]}{\textbf{#1}}

\usepackage{microtype}

\aclfinalcopy 
\def\aclpaperid{26} 

\title{Extending Implicit Discourse Relation Recognition to the PDTB-3}

\author{Li Liang\textsuperscript{1} \qquad Zheng Zhao\textsuperscript{2} \qquad Bonnie Webber\textsuperscript{2} \\
\textsuperscript{1}Dept of Linguistics and English Language, University of Edinburgh\\
\textsuperscript{2}School of Informatics, University of Edinburgh\\
\texttt{L.Liang-7@sms.ed.ac.uk}\\
\texttt{zheng.zhao@ed.ac.uk} \quad \texttt{bonnie@inf.ed.ac.uk}
}

\date{draft - \today}

\begin{document}
\maketitle
\begin{abstract}
The PDTB-3 contains many more implicit discourse relations than the previous PDTB-2. This is in part because implicit relations have now been annotated \textit{within} sentences as well as \textit{between} them. In addition, some now co-occur with explicit discourse relations, instead of standing on their own.
Here we show that while this can complicate the problem of identifying the \textit{location} of implicit discourse relations, it can in turn simplify the problem of identifying their \textit{senses}. We present data to support this claim, as well as methods that can serve as a non-trivial baseline for future state-of-the-art recognizers for implicit discourse relations. 
\end{abstract}

\section{Introduction}

Most readers will be familiar with the PDTB-2 \citep{prasad08}. At the time of its creation, it was the largest public repository of annotated discourse relations (over 43K), including over 18.4K
signalled by explicit discourse connectives (coordinating or subordinating conjunctions,
or discourse adverbials). In the corpus, discourse relations comprise two arguments labelled \argi[Arg1] and \argii[Arg2], with each relation anchored by either
an explicit discourse connective or adjacency. In the latter case, annotators inserted one or more \textit{implicit connectives} to signal the sense(s)
they inferred to hold between the arguments. The size and availability of the PDTB-2 spawned work on
\textit{shallow discourse parsing}, as in the 2015 and 2016 CoNLL shared tasks
\citep{xue-etal-2015-conll,xue-etal-2016-conll}. 

With the release of the PDTB-3\footnote{\url{https://catalog.ldc.upenn.edu/LDC2019T05}}, there are now
$\sim$12.5K additional intra-sentential relations annotated (i.e., relations that lie wholly within the projection of a top-level S-node) and
$\sim$1K additional inter-sentential relations  \citep{pdtb3-manual:2019}. 

Work on \textit{shallow discourse parsing} (including
the CoNLL shared tasks, as well as \cite{bai-zhao-2018-deep,dai-huang-2018-improving,Rutherford-2017,shi-demberg-2017-need}) consistently shows that recognizing and sense labelling implicit discourse relations poses more of a challenge than doing so for explicit discourse relations. Hence, implicit relations
are the focus of the current work.

But there is another reason as well: Work on the
PDTB-2 has assumed (correctly) that non-explicit discourse relations (i.e., implicit relations, \textit{AltLex relations} \citep{prasad-etal-coling2010} and entity relations) only hold between \textit{adjacent sentences} as they did in the PDTB-2, so that a sentence boundary is the only position that needs to be checked for the presence of a non-explicit relation. The difficult problem lay in assigning sense-labels to implicit relations. 

In Section~\ref{sec:ext-mod-pdtb3}, we show that, with the PDTB-3, this is no longer the case because non-explicit relations can hold \textit{within} sentences as well as \textit{between} them.
This in turn motivates a new approach to handle implicit discourse relations in shallow discourse parsing, involving both finding them as well as
identifying their senses (Section~\ref{sec:basic-model}).
After showing that the
sense-distribution of implicit relations \textit{within}
sentences differs from that \textit{between} them
(cf. Section~\ref{sec:sense-dist}), we argue that
one should be able to take advantage of this fact
in sense-labelling these relations.\footnote{Some previous approaches to discourse parsing have also distinguished relations that occur within a sentence from those that occur across sentences \citep{joty-etal-2013-acl,joty-etal-2015-codra}, but it was not felt to be needed in the PDTB-2, where implicit relations only appeared across sentences.}
Section~\ref{sec:inter-intra} describes two
different ways of doing so, along with a way of
dealing with another difference in sense distribution --- that of implicit relations that co-occur with explicit relations and implicit relations that do not.
While the particular methods used here for sense-labelling 
may not advance the state-of-the-art, it is the way we use
them that should deliver a new baseline for recognizing
a fuller range of implicit relations and contribute to the next generation of shallow discourse parsers.\footnote{It would not make sense to have separate processors for explicit discourse relations, as the decision process takes account of the discourse connective, thereby already learning whether the arguments are likely to occur across vs. within sentences.}

\section{Discourse Annotation in PDTB-3}
\label{sec:ext-mod-pdtb3}

Discourse annotation in the PDTB-3 differs from that in the PDTB-2 in two major ways: (1) many more discourse relations are annotated \textit{within} sentences, and (2) there are changes in the sense hierarchy used in annotating them. 
While only the first requires changes to shallow discourse parsing, presenting changes to the senses used in annotating relations will allow us to show differences in the distribution of senses associated with different types of implicit discourse relations.

\subsection{Additional Annotation in PDTB-3}
\label{sec:added-annot}

It was a consequence of the way that the PDTB-2 was annotated, that there were over twice as many discourse relations annotated across sentences than within them. The former were either explicit relations associated with discourse adverbials or sentence-initial coordinating conjunctions\footnote{Despite what people may have been taught, there are over 2100 tokens of sentence-initial "But" in the Penn \textit{WSJ} corpus and over 660 tokens of sentence-initial "And".}, or implicit relations between paragraph-internal adjacent sentences not otherwise linked by a discourse connective.
Within sentences, only annotated were explicit relations associated with subordinating conjunctions, sentence-internal coordinating conjunctions, and discourse adverbials (both of whose
arguments were in the same sentence).  So it should
not be surprising that there were many more inter-sentential relations than intra-sentential relations in the PDTB-2.

In contrast, of the over 13K additional discourse relations annotated in the PDTB-3, over 95\% of them occur
\textit{within} individual sentences. Of the new relations, 5780 are implicit, some standing alone (like the
implicit relations between sentences), with others co-occuring with an explicit discourse relation.
Within a sentence, implicit relations occur at the boundaries of syntactic forms --- for example,
at the boundary between a \textit{free adjunct} and its matrix clause (Ex.~\ref{ex:fa}), or at the boundary between a \textit{to-clause} and its
matrix clause (Ex.~\ref{ex:to}), or between two
punctuation-marked conjuncts (Ex.~\ref{ex:punct}).

\enumsentence{\label{ex:fa}
  \small \argi[Treasury bonds got off to a strong start],
  \argii[advancing modestly during overnight trading on foreign markets]. Conn=\textit{specifically} (\textsc{Arg2-as-detail}) [wsj\_0351]}
\enumsentence{\label{ex:to}
\small
\argi[After a bad start, Treasury bonds were buoyed by a late burst of buying], \argii[to end modestly higher]. Conn=\textit{therefore} (\textsc{Result}) [wsj\_0400]
}
\enumsentence{\label{ex:punct}
\small
Father McKenna moves through the house \argi[praying in Latin], \argii[urging the demon to split]. 
(\textsc{Conjunction}) [wsj\_0413]
}

Because implicit relations within sentences don't
all occur at a single, well-defined position, this
adds to the problems of shallow discourse parsing.

In addition to stand-alone implicits in the PDTB-3, annotators were allowed to indicate implicit relations that co-occur with explicit relations \citep{rohde-etal:iwcs17,rohde-etal:18}, as a way of indicating a relation that
did not derive from the explicit connective, but rather from what the annotator inferred from the arguments themselves, as in Ex.~\ref{ex:multisense1}--\ref{ex:multisense3}:

\enumsentence{\label{ex:multisense1}
\small
We've got to \argi[get out of the Detroit mentality] \argii[and] \conn[Implicit=instead] \argii[be
part of the world mentality], declares Charles M. Jordan,
GM's vice president for design $\ldots$  [wsj\_0956]\\(\textsc{Expansion.Conjunction, Expansion.Substitution.Arg2-as-subst})
}
\enumsentence{\label{ex:multisense2}
\small
$\ldots$ Exxon Corp. \argi[built the plant] \argii[but] \conn[(Implicit=then)]
\argii[closed it in 1985]. [wsj\_1748]\\(\textsc{Comparison.Concession.Arg2-as-denier, Temporal.Asynchronous.Precedence}) 
}
\enumsentence{\label{ex:multisense3}
\small
\ldots which [i.e., the line item veto] would enable him \argi[to kill individual items in a big spending bill] \argii[without] \conn[(Implicit=however)] \argii[having to kill the entire bill]. [wsj\_1133]\\(\textsc{Expansion.Manner.Arg2-as-manner, Comparison.Concession.Arg2-as-denier})
}

In Ex.~\ref{ex:multisense1}, the annotators indicated that they inferred \textsc{Arg2-as-subst} from the pair of arguments
conjoined with \textit{and}. The annotators took \textit{and} itself to convey only that its arguments played the same role with respect to the prior text. It is the arguments themselves that led them to conclude that the second conjunct is meant to substitute for the first.

Similarly, in Ex.~\ref{ex:multisense2}, the annotators indicated that they inferred the temporal relation
\textsc{Precedence} from the pair of arguments conjoined with \textit{but}. The annotators took \textit{but} itself to convey \textsc{Concession}. It is the arguments themselves that led the annotators to conclude that the second conjunct follows the first in time.

Finally, in Ex.~\ref{ex:multisense3}, the annotators indicated that they inferred a \textsc{Concession} relation from the pair of arguments linked by \textit{without}. The annotators took \textit{without} itself (like its positive version \textit{with}) to convey \textsc{Manner}. It is only the arguments that led them to conclude that Arg2 denies an expectation raised by Arg1.

In the PDTB-3, when two relations co-occur, they are
explicitly \textbf{linked} through a shared index. The consequence for shallow discourse parsing is that
explicit relations now need to be checked for co-occurence with an implicit relation.

\subsection{Changes to the Sense Hierarchy}

The sense hierarchy used in annotating the PDTB-3
differs from that used in annotating the PDTB-2 in three ways:
\begin{enumerate}
    \item Rare and/or difficult to annotate senses were dropped, as with the different types of conditional senses;
    \item Sense relations at Level-3 now only encode \textit{directionality} --- for example, distinguishing \textsc{Arg1-as-subst} (Ex.~\ref{ex:subst-level3-arg1}) from \textsc{Arg2-as-subst} (Ex.~\ref{ex:subst-level3-arg2})
    \item New senses were added that were found to be needed for annotating relations within sentences.
\end{enumerate}
\enumsentence{\label{ex:subst-level3-arg1}
\small \textsc{Arg1-as-subst:} \conn[instead of] \argii[featuring a major East Coast team against a West Coast team], \argi[it pitted the Los Angeles Dodgers against the losing Oakland A's] [wsj\_0443]}
\enumsentence{\label{ex:subst-level3-arg2}
\small He \argi[could develop the beach through a trust],
\argii[but] \conn[instead] \argii[is trying have his grandson become a naturalized
Mexican so his family gains direct control]. [wsj\_0300]}

More about the senses used in annotating the PDTB-3 can be found in \citet{pdtb3-manual:2019}.
Senses are relevant to this discussion of
implicit relations in shallow discourse parsing because (as set out in Section~\ref{sec:sense-dist}) implicit relations have been found to have
different sense distributions depending on where
they occur.

\subsection{Stand-off annotation in the PDTB-3}

Both the PDTB-2 and PDTB-3 use stand-off annotation.  What is relevant with respect to the experiments we report here, is what information is explicit in the annotation, as opposed to having to be computed. This information includes (1) the type of the relation (Explicit, Implicit, AltLex, AltLexC, Entity, Hypophora, NoRel); (2) the byte spans of the two arguments of the relation; and (3) the explicit index (aka \textit{link}) of relations that co-occur by virtue of sharing the same or nearly the same arguments.
The full field structure of discourse relations is set out in Section 8 of \citet{pdtb3-manual:2019}.
What has to be recovered from the argument spans and the span of the projection of the top node in each sentence-level parse tree is whether a relation occurs wholely within a single sentence or involves multiple sentences.

\section{Basic Model Architecture}
\label{sec:basic-model}

The sense classifiers for implicit relations used in this paper are based on
a Basic Model whose properties reflect consideration of data size and the interaction between lexical information and structural information.
(A full description of the Basic Model is given in Appendix~A.)

\begin{figure*}
    \centering
    \includegraphics[scale=0.5]{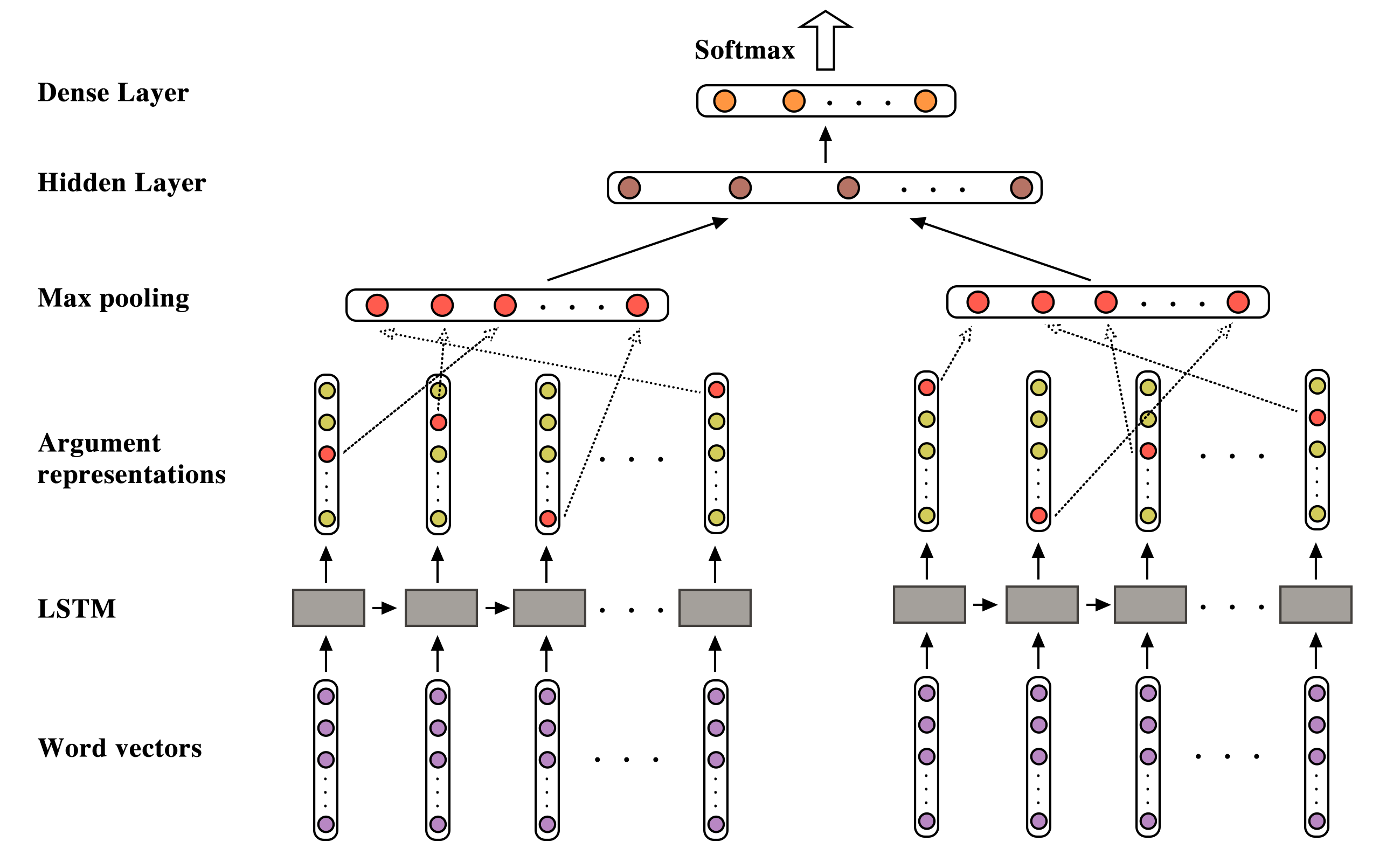}
    \caption{The overall model architecture for implicit sense classification}
    \label{fig:basic-model}
\end{figure*}

The architecture of Basic Model is shown in Figure \ref{fig:basic-model}. It consists of two LSTMs \citep{hochreiter1997long} and max-pooling layers, a hidden layer, a dense layer, and a softmax layer. Inputs to the model consist of pairs of discourse arguments, each represented as a sequence of word vectors. The output is a probability distribution of the senses between the discourse argument spans.
The two sequences of word vectors are encoded by LSTMs in order to capture positional information within the sequential structure. 
Max-pooling on the output of the LSTMs is used to compose meaning and reduce parameters for the model, as it has been proven effective in \citet{conneau-etal-2017-supervised}.
Modeling the interaction between discourse arguments follows \citet{rutherford-xue-2016-robust}, who argue that discourse relations can only be determined by jointly analyzing the arguments. In addition, \citet{Rutherford-2017} observed the influence of different configurations on the performance of the model for the implicit sense classification task, suggesting an interaction between the lexical information in word vectors and the structural information encoded in the model itself. We follow them in adopting a 300-dimension word2vec \citep{10.5555/2999792.2999959} word embedding and hidden size of 100 for the Basic Model.

\section{Differences in the distribution of sense relations}
\label{sec:sense-dist}

To argue for separating the recognition of intra-sentential implicits from inter-sentential implicits, and the recognition
of linked implicits from stand-alone implicits,
we show how their sense distributions are different.

Table \ref{table:inter-intra} compares the distribution of inter-sentential and intra-sentential implicit relations with respect to the PDTB-3's Level-2 sense labels, along with the proportion of each label to the total inter-sentential and intra-sentential implicit relations. Besides differences in frequency --- for example, relations expressing \textsc{Purpose} constitute 21.76\%
of intra-sentential implicit relations, while only 0.12\% of inter-sentential implicits, while relations expressing \textsc{Instantiation} constitute 8.89\% of inter-sentential implicits, while only 1.4\% of intra-sentential implicits ---
the senses of inter-sentential implicits are more
unequally distributed. That is, three senses --- \textsc{Contingency.Cause}, \textsc{Expansion.Conjunction} and \textsc{Level-of-detail} cover 67.08\% of the inter-sentential implicits. In contrast, except for \textsc{Contingency.Cause} and \textsc{Purpose}, most of the other intra-sentential implicits are more evenly distributed. As often happens with training on an imbalanced distribution, the unequal distribution of inter-sentential relations can lead the model to predict the majority class, ignoring minority classes.

\begin{table*}[t]
\begin{small}
    \centering
    \begin{tabular}{|c|l|l|l|}
\hline
& & inter-sentential & intra-sentential \\
\hline
{Comparison}& Concession& 1355\hfill(8.70\%)& 136\hfill(2.19\%)\\
& Concession+SpeechAct& 7\hfill(0.04\%)& 3\hfill(0.05\%)\\
& Contrast& 700\hfill(4.50\%)& 156\hfill(2.51\%)\\
& Similarity& 14\hfill(0.09\%)& 14\hfill(0.23\%)\\
\hline
{Contingency}& Cause& 4153\hfill(26.67\%)& 1613\hfill(25.97\%)\\
& Cause+SpeechAct& 21\hfill(0.13\%)& 1\hfill(0.02\%)\\
& Cause+Belief& 105\hfill(0.67\%)& 94\hfill(1.51\%)\\
& Condition& 1\hfill(0.01\%)& 198\hfill(3.19\%)\\
& Condition+SpeechAct& 1\hfill(0.01\%)& 1\hfill(0.02\%)\\
& Purpose& 19\hfill(0.12\%)& 1351\hfill(21.76\%)\\
\hline
{Expansion}& Conjunction& 3648\hfill(23.43\%)& 733\hfill(11.80\%)\\
& Disjunction& 9\hfill(0.06\%)& 21\hfill(0.34\%)\\
& Equivalence& 286\hfill(1.84\%)& 48\hfill(0.77\%)\\
& Exception& 4\hfill(0.03\%)& 1\hfill(0.02\%)\\
& Instantiation& 1385\hfill(8.89\%)& 87\hfill(1.40\%)\\
& Level-of-detail& 2644\hfill(16.98\%)& 589\hfill(9.48\%)\\
& Manner& 4\hfill(0.03\%)& 223\hfill(3.59\%)\\
& Substitution& 221\hfill(1.42\%)& 145\hfill(2.33\%)\\
\hline
{Temporal}& Asynchronous& 647\hfill(4.15\%)& 608\hfill(9.79\%)\\
& Synchronous& 348\hfill(2.23\%)& 188\hfill(3.03\%)\\
\hline
total &  & 15572 & 6210 \\
\hline
    \end{tabular}
    \caption{Distribution of inter-sentential/intra-sentential implicit relations among Level 2 labels and the proportion of each label with respect to inter-sentential/intra-sentential implicit relations}
    \label{table:inter-intra}
    \end{small}
\end{table*}

As for the 1753 implicits that co-occur with explicit relations, Table~\ref{table:linked-SA} shows that their sense distribution differs sharply from that of stand-alone implicit relations. For example, over 70\% convey either \textsc{Cause} or \textsc{Asynchronous},
while this holds of only 28.7\% of stand-alone implicit relations. As such, linked implicits should be more predictable than stand-alone implicit relations.
\begin{table*}[t]
\begin{small}
    \centering
    \begin{tabular}{|c|l|l|l|}
 \hline
& & stand-alone & linked \\
\hline
{Comparison}& Concession & 1401\hfill(6.99\%)& 90\hfill(5.13\%)\\
& Concession+SpeechAct  & 10\hfill(0.05\%) & 0\hfill(0.00\%)\\
& Contrast & 795\hfill(3.97\%)& 61\hfill(3.48\%)\\
& Similarity & 18\hfill(0.09\%)&10\hfill(0.57\%)\\
\hline
{Contingency}& Cause & 4943~\hfill(24.68\%) & 823~\hfill(46.95\%)\\
& Cause+SpeechAct  & 22\hfill(0.11\%)& 0\hfill(0.00\%)\\
& Cause+Belief & 164\hfill(0.82\%)& 35 \hfill(2.00\%)\\
& Condition & 199\hfill(0.99\%)&0 \hfill(0.00\%)\\
& Condition+SpeechAct & 2\hfill(0.01\%)& 0\hfill(0.00\%) \\
& Purpose& 1367\hfill(6.83\%)&3\hfill(0.17\%)\\
\hline
{Expansion} & Conjunction& 4360~\hfill(21.77\%)& 21\hfill(1.20\%) \\
& Disjunction & 30\hfill(0.15\%)& 0\hfill(0.00\%)\\
& Equivalence& 326\hfill(16.28\%)& 8\hfill(0.46\%)\\
& Exception& 4\hfill(0.02\%)& 1\hfill(0.06\%)\\
& Instantiation& 1456\hfill(7.27\%)& 16 \hfill(0.91\%)\\
& Level-of-detail& 3172~\hfill(15.84\%)& 61 \hfill(3.48\%)\\
& Manner & 173\hfill(0.86\%)& 54\hfill(3.08\%)\\
& Substitution & 276\hfill(1.38\%)& 90\hfill(5.13\%)\\
\hline
{Temporal}& Asynchronous& 800\hfill(3.99\%)& 455~\hfill(25.96\%)\\
& Synchronous& 511\hfill(2.55\%)& 25\hfill(1.43\%)\\
\hline
total &  & 20029 & 1753 \\
\hline
    \end{tabular}
    \caption{Distribution of linked and stand-alone implicit relations among Level 2 labels and the proportion of each label with respect to the total linked/stand-alone implicit relations}
    \label{table:linked-SA}
\end{small}
\end{table*}

\section{Inter- and intra-sentential Implicits}
\label{sec:inter-intra}

Differences in the distribution of implicit relations
\textit{within} sentences and \textit{across} sentences
suggest that we exploit this difference in sense-labelling implicit relations. In this section, we first assume that we know where implicit relations are located within a sentence, so that we can simply consider their arguments. We then present work we have done towards relaxing this assumption.

\paragraph{Task 1: Consider the location of implicit relations in classification.} There are different ways to take the location of implicit relations into consideration. Here we present two models, \textbf{Model 1} (Section~\ref{subsec:Model 1}) and \textbf{Model 2} (Section~\ref{subsec:Model 2}), 
both based on the basic model architecture described in Section~\ref{sec:basic-model}. We compare them with the \textbf{Basic Model}, which uses the same classifier on all tokens. We compare their performance not just using the standard training-development-test split, where the ratio of inter- to intra-sentential implicits in the training set, WSJ section 2-21, is 12787:5014. In addition, we follow \citet{shi-demberg-2017-need}, who argue that evaluation through cross-validation is more predictive, given the wide variation in texts that appear in different sections of the Penn \textit{Wall Street Journal} corpus. The average ratio of inter- to intra-sentential implicits in training sets of cross-validation is 12747:4992. The scores of 3 models are weighted by the proportion of inter- and intra-sentential tokens in the test set.

\paragraph{Task 2: Identify the location of implicit relations.} To reduce the dependency on the gold standard annotations of where implicit discourse relations hold within sentences, two recognizers to identify implicit relations and find argument spans are provided. The first recognizer (Section~\ref{sec:zheng}) takes syntactic features to identify sentences that contain intra-sentential relations. The second recognizer (Section~\ref{sec:5.5}) exploits the properties that some explicit relations are linked with implicit relations, checking the explicit relations for co-occurrence with implicit relations to obtain the shared arguments. 
 
 \begin{table*}[t]
 \begin{small}
    \centering
    \begin{tabular}{|c|c|c|c|c|}
    \hline
& \multicolumn{3}{c|}{Main evaluation metric} & 
Cross \\
\cline{2-4}
 & inter-sentential & intra-sentential & overall & validation\\
\hline
Basic model & 35.791 & 47.154 & 38.608 & 41.463\\
Model 1 & 34.973 & 56.666 & 40.222 & \textbf{43.418} \\
Model 2 & 37.701 & 50.410 & \textbf{40.827} & 42.174\\
    \hline
    \end{tabular}
    \caption{$F_1$ scores of the different models on inter-sentential and intra-sentential implicit relation at Level 2.}
\label{tab:compare-models}
\end{small}
\end{table*}

 \subsection{Basic Model}
 \label{subsec:basic model}
 
\begin{figure*}[t]
 \centering
    \includegraphics[scale=0.53]{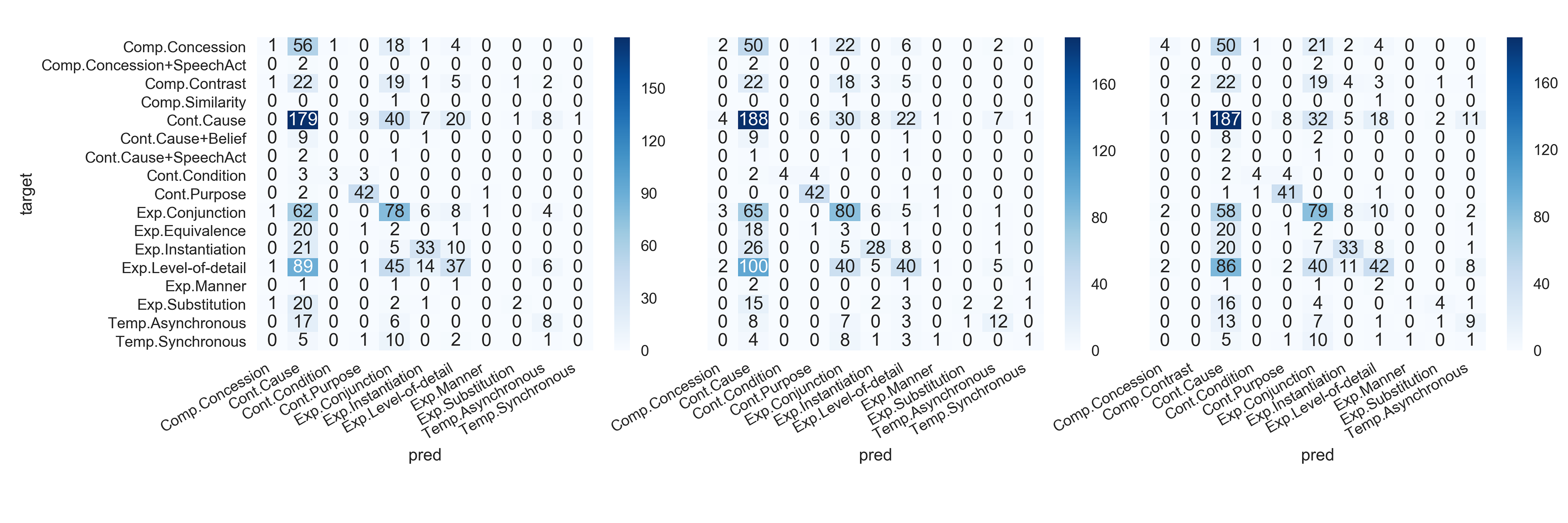}
    \caption{Confusion matrix of the Basic Model, Model 1 and Model 2}
    \label{fig:conf-matrix}
\end{figure*}

 The Basic Model uses the same classifier on all tokens.
 Since we know which tokens are inter-sentential
 and which are intra-sentential, we can compare how well
 the Basic Model does on each. To compute the $F_1$ scores for the overall performance of the model, the scores of the model are combined, weighted by the proportion of inter- or intra-sentential tokens in the test set. This is shown on the first line of Table~\ref{tab:compare-models}, elaborated in
 the confusion matrix shown in Figure~\ref{fig:conf-matrix}. A Chi-squared test on the results show the performance of the Basic Model appears to depend to a statistically significant extent on whether the sense appears inter- or intra-sententially (p=1.50e-03). 
 
\subsection{Model 1}
\label{subsec:Model 1}
\noindent\textbf{Model architecture:}
The idea behind Model 1 is to separate the classification task into intra-sentential and inter-sentential implicit sense classification, with separate classifiers for each. The model architecture and configuration of each classifier are the same as in the Basic Model (Section~\ref{sec:basic-model}). We expect each classifier to capture different sense distributions of intra-sentential or inter-sentential implicits.


\vspace{.2cm}
\noindent\textbf{Training and evaluation:} Based on their argument spans and the spans associated with each sentence in a file, tokens can be labeled as inter-sentential or intra-sentential. For the standard training-development-test framework, the tokens are
allocated into separate inter-sentential/intra-sentential training, development, and test sets. The inter-sentential training set is used in training the inter-sentential implicit sense classifier, and similarly for intra-sentential classification. Test set tokens labeled as inter-sentential or intra-sentential are fed into the appropriate classifier. 

\vspace{.2cm}
\noindent\textbf{Results:} 
The second line of Table~\ref{tab:compare-models} presents $F_1$ scores for Model 1 evaluated on the main evaluation test set and by cross-validation. It shows that Model 1 improves on the Basic Model in predicting intra-sentential implicit relations. The performance of the model significantly depends on the location of relations (p = 2.41e-09). The confusion matrix for Model~1\footnote{combining results of the inter-sentential and intra-sentential classifiers} (cf. Figure~\ref{fig:conf-matrix}) shows that labels with a relatively larger sample size in each set are predicted more often, including \textsc{Contingency.Purpose} (frequent in intra-sentential implicits), \textsc{Expansion.Conjunction} (frequent in inter-sentential implicits) and \textsc{Contingency.Cause} (frequent in both). The confusion matrix also shows that less frequent senses are confused with these frequent labels more often. Model 1 also reduces the ignorance problem of the Basic Model,
in that it correctly classifies some samples into \textsc{Temporal.Synchronous}, which is a label ignored by the basic model. 

\subsection{Model 2}
\label{subsec:Model 2}
\noindent\textbf{Model architecture:} Model~2 treats being inter-sentential or intra-sentential as a single binary feature. Model~2 is created by modifying the Basic Model to include this feature after obtaining the combined representations of the two arguments. We concatenate the binary feature $f_S$ with the output of the dense layer before applying the softmax function, expecting it to affect the final prediction.

\vspace{.2cm}
\noindent\textbf{Training and evaluation:} The data selection follows the standard and cross-validation data split process. The evaluation assumes that each token in the test set has been given an inter-sentential or intra-sentential feature. The scores are computed following the general process as the basic model.


\vspace{.2cm}
\noindent\textbf{Results:} The third line of
Table~\ref{tab:compare-models} shows that Model 2 improves over the Basic Model with respect to both inter- and intra-sentential implicit sense prediction, though the performance of the model still has a statistically significant dependence on the location of relations (p = 4.53e-04). The improvement of Model 2 on intra-sentential labels is not as dramatic as Model 1. Compared to the previous model, Model 2 doesn't sharpen its focus on those frequent labels in inter- or intra-sentential sets. Instead, the integrated feature in the representations distributes the benefits on the prediction ability of different labels more evenly. In addition, the confusion matrix in Figure \ref{fig:conf-matrix} shows that Model 2 reduces the confusion between \textsc{Instantiation} and \textsc{Level-of-detail}, which \citet{scholman-demberg17} have hightlighted as a common source of confusion.
The confusion matrix for Model 2 also shows some attention to less frequent labels such as 
\textsc{Comparison.Contrast}, which are not predicted in either the Basic Model or Model 1.

\subsection{Towards finding implicits within sentences}
\label{sec:zheng}

The results presented above reflect ``gold knowledge''
of where implicit discourse relations hold within sentences.  But in truth, their locations need
to be identified before (or jointly with) labelling
their senses. We have viewed this as a two-step process: Recognizing sentences that contain at least one implicit intra-sentential relation, and then recognizing the arguments to each relation. The first step has been implemented using a recognizer that takes a linearized parse tree of a sentences as the input. The second step is future work.

\vspace{.2cm}
\noindent\textbf{Model architecture:} Similar to the Basic Model, inputs are represented as a sequence of word vectors, and word embeddings are initialized using pre-trained fastText \citep{bojanowski-etal-2017-enriching} vectors (16B tokens). These vectors are fed to a BiLSTM whose outputs are then fed to a linear layer to produce a binary label, indicating the existence of at least one implicit intra-sentential relation.  Word embeddings are set to 200, hidden dimensions, to 256, and vocabulary size, to 25k. 

\vspace{.2cm}
\noindent\textbf{Training and evaluation:} To train our recognizer, we first created a dataset of triplets comprising a sentence from PDTB-3, its corresponding parse tree, and a binary label. We obtain the parse trees from the Penn TreeBank (PTB -- \citealt{marcus-etal-1993-building}) and set the binary label to 1 if there exist at least one implicit or AltLex relation in that sentence. For example, the sentence in Ex.~\ref{ex:dataset_implicit} is labelled 1, while that in
Ex.~\ref{ex:dataset_explicit} is labelled 0. 
\enumsentence{\label{ex:dataset_implicit}
  \small MARKET MOVES, these managers don't. 
  \\
  ( ( S-HLN ( S ( NP-SBJ ( NN  MARKET ) ) ( VP ( VBZ MOVES ) ) ) ( , , ) ( S ( NP-SBJ ( DT  these ) ( NNS  managers ) ) ( VP ( VBP  do ) ( RB  n't ) ( VP ( -NONE-  *?* ) ) ) ) (  . . ) ) )
[wsj\_1825]
}

\enumsentence{\label{ex:dataset_explicit}
  \small Oil-tool prices are even edging up. 
  \\
  ( ( S ( NP-SBJ ( NN  Oil-tool ) ( NNS  prices ) ) ( VP ( VBP  are ) ( ADVP ( RB  even ) ) ( VP ( VBG  edging ) ( ADVP-DIR ( RP  up ) ) ) ) (  . . ) ) )
[wsj\_0725]
}
Intra-sentential AltLex relations are included here because they are
simply Implicit relations whose alternative lexicalization reliably signals its sense ---
for example, the phrases "resulting in",
"avoiding", and "contributing to" are all taken to be alternative lexicalizations that reliably signal \textsc{Result}.  This is not true of the earlier Examples~\ref{ex:fa}--\ref{ex:punct}, which are classed as Implicits. On the other hand, we do not label ``linked'' implicit relations as 1 because the visible evidence is an explicit connective signalling an explicit relation, and we don't want that to be taken \textit{per se} as evidence for an implicit relation. For recognizing linked implicits, we have built a separate model which will be discussed in Section~\ref{sec:5.5}. 

Our training used the Adam optimizer \citep{KingmaB14} with a learning rate of 1e-4. We randomly split the dataset into training (60\%), development (20\%) and test (20\%). To understand what happens if ``gold parse trees'' are not used, we also created variants of the dataset using parse trees from the widely used Berkeley parser \citep{kitaev-klein-2018-constituency} and Stanford parser \citep{manning2014stanford}. 

\vspace{.2cm}
\noindent\textbf{Results:} As the dataset is heavily imbalanced, we also added a simple baseline which predicts the most frequent label. Test set results of the recognizer on the three datasets are presented in Table \ref{tab:intraS-rel-recognizer}. Even though the baseline achieved an accuracy of $\sim$0.9, it doesn't convey any useful information, as it labels all instances as 0. We can observe that the model with gold Penn TreeBank parse trees obtain the best performance, followed by the Berkeley parser. Stanford parse trees result in worst performance. Examining these trees led us to conclude that, while the Stanford parser does well for basic syntactic structures, which are the most common, it has trouble with challenging structures such as those associated with conjunction. An example is provided in Ex.~\ref{ex:stanford_parse}. Here, ``steps'' has been incorrectly labelled NNS, when it is actually a VBZ, heading the second conjunct. If there were only two conjuncts, explicitly conjoined with ``and'', the sentence would not contain an implicit relation. With three conjuncts, however, the first two would normally be \textit{comma-conjoined}, with the discourse relation between them taken to be implicit. But the error in PoS-tagging has eliminated evidence of a second conjunct, with an implicit discourse relation to the first conjunct. Errors in PoS-tagging and mis-parsing associated with rare constructions, means that the  accuracy is lower than that of the Berkeley parser. However, as Precision, Recall, and F$_1$ are measured for 1 labels, these metrics are more adversely affected when compared to those of the Berkeley parser.  

\enumsentence{\label{ex:stanford_parse}
  \small With three minutes left on the clock, Mr. Aikman takes the snap, steps back and fires a 21-yard pass -- straight into the hands of an Atlanta defensive back. 
  \\
  \smallskip
  \\
  IN CD NNS VBD IN DT NN , NNP NNP VBZ DT NN , NNS RB CC VBZ DT JJ NN : RB IN DT NNS IN DT NNP NN RB . \\
  \smallskip
  \\
  ((S    
    (SBAR (IN With)
      (S
        (NP (CD three) (NNS minutes))
        (VP (VBD left)
          (PP (IN on)
            (NP (DT the) (NN clock))))))
    (, ,)
    (NP (NNP Mr.) (NNP Aikman))
    (VP
      (VP (VBZ takes)
        (NP
          (NP (DT the) (NN snap))
          (, ,)
          (NP (NNS steps)))
        (ADVP (RB back)))
      (CC and)
      (VP (VBZ fires)
        (NP (DT a) (JJ 21-yard) (NN pass))
        (: --)
        (PP (RB straight) (IN into)
          (NP
            (NP (DT the) (NNS hands))
            (PP (IN of)
              (NP (DT an) (NNP Atlanta) (NN defensive))))))
      (ADVP (RB back)))
    (. .)))
[wsj\_1411]
}

\begin{table}[]
\begin{small}
    \centering
    \begin{tabular}{|c|c|c|c|c|}
    \hline
      Parse trees & Accuracy & Precision & Recall & $F_1$\\
      \hline
      Baseline & 0.9028 & 0 & 0 & 0\\
      Gold  & 0.9617 & 0.7799 & 0.8968 & 0.8343 \\
    Berkeley & 0.9473 & 0.7814 & 0.6334 & 0.6997\\
    Stanford & 0.9349 & 0.7153 & 0.5537 & 0.6242\\
    
    \hline
    \end{tabular}
    \caption{Results on task of identifying sentences that contain at least one intra-sentential relation, comparing gold parse trees from the PTB with the parse trees output by the Berkeley parser and by the Stanford parser. Baseline refers to the model that predicts the most frequent label.}
    \label{tab:intraS-rel-recognizer}
    \end{small}
\end{table}

\subsection{Recognizing ``linked'' implicit relations}
\label{sec:5.5}
As noted in Section~\ref{sec:added-annot}, implicit
relations can co-occur with explicit relations. While
the location of such implicits is not identified by the recognizer described in Section~\ref{sec:zheng}, we
actually know the location of their arguments, because co-occurring (aka ``linked'') relations share their argument spans. Hence, recognizing explicit relations linked with implicit ones means that we also obtain argument spans of these implicits. Here we
describe a first attempt to automatically discriminate explicit relations linked with implicit relations from ones that are not so linked. It comprises two steps: extracting sentences that contain explicit relations as our datasets, and then recognizing the ones linked with implicit relations.

\vspace{.2cm}
\noindent\textbf{Model architecture:} To detect linked implicit relations from explicit relations, we use a naive Bayes classifier --- specifically, the one provided in NLTK \citep{bird-loper-2004-nltk}. Production rules are selected as input feature as it has been proven notably effective in feature-based implicit discourse relation recognition task among different features \citep{park-cardie-2012-improving}. Models trained in Task 1 will be adopted for linked sense classification.

\vspace{.2cm}
\noindent\textbf{Training and evaluation:} We follow the standard split to select the training and test set. Each token in the training set consists of \argi[Arg1], connective and \argii[Arg2], and are parsed to extract syntactic productions used in parent-child nodes in the argument parse trees. The 100 most-frequent production rules are used to build a feature dictionary for input. A production rule feature is labeled as 1 in the dictionary if it appears in the parse tree of the token, otherwise it will be 0. The linked/stand-alone label is determined by whether the explicit relation shares the same index value with an implicit relation. The recognizer is evaluated by how well it distinguishes explicit relations that have a linked implicit relation from ones that don’t. Classifiers are evaluated on the recognized implicit relations.

\begin{table}[]
\begin{small}
    \centering
    \begin{tabular}{|l|c|c|c|c|}
    \hline
& Precision & Recall & $F_1$ & Proportion\\
\hline
stand-alone & 0.951 &  0.905 &  0.928 & 93.67\%\\
linked & 0.193  & 0.329 &  0.243 & 6.33\%\\
    \hline
    \end{tabular}
    \caption{Precision, Recall and $F_1$ scores of linked/stand-alone labels predicted by the recognizer using main evaluation metrics and their proportion in test data. }
    \label{tab:linked-recognizer}
    \end{small}
\end{table}

\vspace{.2cm}
\noindent\textbf{Results:} The low Recall for linked relations in Table \ref{tab:linked-recognizer} shows that the recognizer performs better on predicting stand-alone relations, which are a majority of the data. Linked implicits in the test set (WSJ Section 23) are mostly linked to conjoined clauses or conjoined VPs, and are signaled by implicit connective like ``and'' (81.08\%) or ``but'' or an adverbial. Most correctly recognized relations are VPs conjoined with ``and''. All the recognized linked implicit relations are found intra-sentential. We adopt the intra-sentential classifier in Model 1 and the Basic Model to test the classifier based on the recognized results. The intra-sentential classifier achieves an $F_1$ score of 75, compared with 68.182 using the Basic Model. This again emphasizes that knowing the location of implicit discourse relation would benefit sense identification.

\section{Conclusion and future work}

We have shown that recognizing implicit discourse relations as annotated in the PDTB-3 now requires finding them, as well as figuring out what sense relation(s) holds between
the arguments. However, we have also shown that the latter task is simplified by differences in the sense distribution of different implicit relations.
We still have to develop a way of recognizing precisely where implicit relations hold in those sentences that can be identified as containing them, and a more accurate approach to sense labelling implicit relations that co-occur with explicit ones.
We are also interested in whether these different sense distributions hold in other news corpora and other genres. While it is likely not the case that all languages show the same difference in the sense distribution of discourse relations, we would not be surprised if the discourse relations realized within sentences differed from those realized across sentences.
In conclusion, we hope that the current effort will contribute to future work on shallow discourse parsing as
annotated in the PDTB-3.

\section*{Acknowledgments}
We would like to thank the anonymous reviewers for their valuable comments. We would also like to thank Annie Louis for her contributions to the work on recognizing the presence of sentence-internal implicit discourse relations.

\bibliographystyle{acl_natbib}
\bibliography{codi2020}

\newpage

\appendix

\counterwithin{figure}{section}
\counterwithin{table}{section}

\section{Specifics of the Basic Model}

\label{app:basic-model}

Here we describe the basic model architecture for implicit relation sense classification in PDTB-3. The configuration for the model is chosen based on consideration of data size and the interaction between lexical information and structural information. A further analysis on the predictive performance of the basic model on each labels is provided as well.

\subsection{Model architecture}
\begin{figure*}
    \centering
    \includegraphics[scale=0.65]{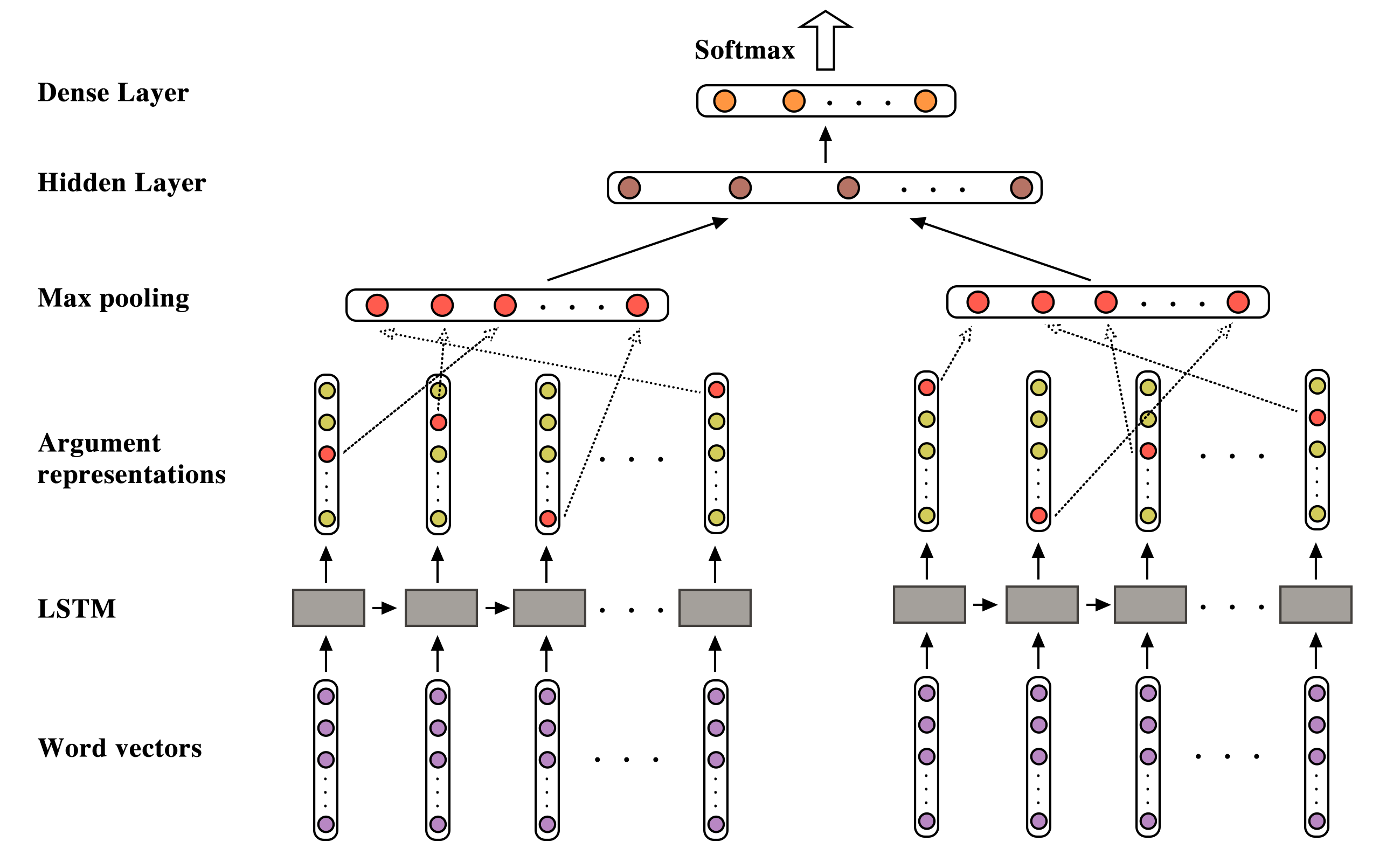}
    \caption{The overall model architecture for implicit sense classification}
    \label{fig:4.1}
\end{figure*}

 Figure \ref{fig:4.1} illustrates the overall model architecture of the neural implicit sense classifier that consists of two LSTM and max-pooling layers, a hidden layer, a dense layer, and a softmax layer. The input for the model is the discourse argument pairs with additional labels\footnote{These labels are not used in the basic model described in this work, but serve for statistical tests and further experiments.}, and the output is a probability distribution of the senses between the discourse argument spans.

\vspace{.2cm}
\noindent\textbf{Word vectors:} In our model, arguments $Arg1$ and $Arg2$ are viewed as two sequences of word vectors with length of $n_1$ and $n_2$. Word vectors for the word in arguments are taken from word embeddings. 
\begin{equation}
    Arg1: [x^1_{1}, x^1_{2}, ...,x^1_{n_1}]
  \end{equation}  
 \begin{equation}
    Arg2: [x^2_{1}, x^2_{2}, ...,x^2_{n_2}]
\end{equation}

\vspace{.2cm}
\noindent\textbf{Argument representations:} The two sequences of word vectors are encoded by LSTM respectively. 
The hidden states $H_{Arg1}$ and $H_{Arg2}$ of LSTM are taken. The max-pooling function is employed to compose meaning in the hidden states and reduce parameters for the model, as it has been proven effective in \cite{conneau-etal-2017-supervised}. As shown in eq. \ref{eq:max-pooling}, it will select the maximum value along the sequence at each dimension of the hidden states. $a^1_j$($a^2_j$) represents a maximum value from all the values in a sequence with length of $n_1$($n_2$) at dimension $j$ of the hidden states $H_{Arg1}$ ($H_{Arg2}$). By concatenating the output of max-pooling function, we have abstract representations $A_{Arg1}$ and $A_{Arg2}$ of arguments $Arg1$ and $Arg2$ individually.
\begin{equation}
    H_{Arg1} = [h^1_{1}, h^1_{2}, ...,h^1_{n_1}]
\end{equation} 

\begin{equation}
    H_{Arg2} = [h^2_{1}, h^2_{2}, ...,h^2_{n_2}]
\end{equation}

\begin{equation}
a^1_j = \max_{k\in n_1}(H_{{Arg2}_{{j}_k}})
\end{equation}

\begin{equation}
a^2_j = \max_{k\in n_2}(H_{{Arg1}_{{j}_k}})
\label{eq:max-pooling}
\end{equation}

\begin{equation}
A_{Arg1}= [a^1_{1}, a^1_{2}, ...,a^1_{hidden\_size}]
\end{equation}

\begin{equation}
A_{Arg2} = [a^2_{1}, a^2_{2}, ...,a^2_{hidden\_size}]
\end{equation}

\vspace{.2cm}
\noindent\textbf{Inter-argument interaction modeling:} The modeling of the interaction between two discourse argument representations follows \cite{rutherford-xue-2016-robust}, which argues that discourse relations can only be determined by jointly analyzing the arguments. In our model, argument representations $A_{Arg1}$ and $A_{Arg2}$ are weighted by $W_1$ and $W_2$ separately. The combination of the weighted argument representations is then transformed non-linearly with $tanh$ function in the first hidden layer $H_{hid}$. It is then fed into a dense layer $H_{dense}$\footnote{The default size of the dense layer is $hidden\_size//5$.}. Finally, we predict the discourse relation sense using a softmax function.
\begin{equation}
    H_{hid} = tanh(W_1\cdot A_{Arg1}+W_2\cdot A_{Arg2}+b_{hid})
\end{equation}
\begin{equation}
    H_{dense} = tanh(W_{dense}\cdot H_1+b_{dense})
\label{eq:dense}
\end{equation}
\begin{equation}
    output = softmax(W_{output}\cdot H_{dense}+b_{output})
\end{equation}

\vspace{-.2cm}
\subsection{Configuration}
\noindent\textbf{Implementation:} The model is implemented with PyTorch. The cost function is the standard cross-entropy loss function and Adam optimizer with an initial learning rate of 0.001 and a batch size of 32. We determine convergence if the performance of the model on the development set does not improve after more than 3 epochs.

One problem that challenges the training of the model is the limitation on the size of the data. We introduce other resources to overcome it and adopt different techniques to avoid overfitting. Word vectors are directly taken from Word2vec embeddings \citep{W2v} trained with the skip-gram algorithm on Brown corpus, and are fixed during training. To avoid overfitting, we apply a 0.25 dropout ratio to the input of the LSTM layer. Batch normalization is added to normalize the activation between the hidden layer and the dense layer to accelerate the training speed and further prevent overfitting with regularization. 

\vspace{.2cm}
\noindent\textbf{Hyperparameter Settings:} \cite{Rutherford-2017} observed the influence of different configurations on the performance of the model for the implicit sense classification task, suggesting an interaction between the lexical information in word vectors and the structural information encoded in the model itself. To determine the configuration for our model, we trained our model with different combinations of the dimension of word embedding (50, 300) and hidden size (50, 100), and evaluate it on Level 2 labels on the WSJ section 23. Table \ref{table:4.1} presents the performance of the model with different configurations. The baseline is Most Frequent Sense heuristic, using the most frequent sense \textsc{Contingency.Cause} in the training data for each target. Our result is in line with their finding of sequential LSTM model, showing larger hidden size 100 is effective when it is accompanied with 300-dimension word embedding. Based on the performance on Level 2 labels, we choose 300-dimension Word2vec word embedding and hidden size 100 as our configuration for the Basic Model. 

\begin{table*}[]
\begin{small}
    \centering
    \begin{tabular}{|c|c|c|c|}
    \hline
 & embedding size & hidden size & $F_1$ \\
    \hline
{Our model} & 50 & 50 & 36.492 \\
& 50 & 100 & 37.097 \\ 
& 300 & 50 & 37.601 \\
& 300 & 100 & \textbf{38.608} \\
\hline
Baseline & - & - & 28.024 \\
    \hline
    \end{tabular}
    \caption{$F_1$ scores of the model with different configurations on predicting Level 2 sense labels}
    \label{table:4.1}
\end{small}
\end{table*}

Our model scores 34.778 at Level 3 (31-way classification). Using cross-validation, our model obtains 41.463 at Level 2. 

\subsection{Discussion}
It is worth examining the performance of the model on each Level 2 label individually.  Table \ref{table:b} displays the precision, recall and $F_1$ scores of each label along with its proportion in the test data. 

\begin{table*}[]
\begin{small}
    \centering
    \begin{tabular}{|l|c|c|c|c|}
    \hline
& Precision & Recall &  $F_1$  & Proportion \\
\hline
Comparison.Concession & 20.000 & 1.235 & 2.326 & 8.17\%\\ 
Comparison.Concession+SpeechAct &  0.000 & 0.000 & 0.000 &0.20\%\\ 
Comparison.Contrast & 0.000 &  0.000 & 0.000 & 5.14\%\\
Comparison.Similarity&0.000&0.000&0.000 & 0.10\%\\
Contingency.Cause&35.098&67.547&46.194 & 26.71\%\\
Contingency.Cause+Belief&0.000&0.000&0.000&1.01\%\\
Contingency.Cause+SpeechAct&0.000&0.000&0.000&0.30\%\\
Contingency.Condition&75.000&33.333&46.154&0.91\%\\
Contingency.Purpose&73.684&93.333&82.353&4.54\%\\
Expansion.Conjunction&34.211&48.750&40.206&16.13\%\\
Expansion.Equivalence&0.000&0.000&0.000&2.42\%\\
Expansion.Instantiation&51.562&47.826&49.624&6.96\%\\
Expansion.Level-of-detail&42.045&19.171&26.335 &19.46\%\\
Expansion.Manner&0.000&0.000&0.000&0.30\%\\ 
Expansion.Substitution&50.000&7.692&13.333&2.62\%\\
Temporal.Asynchronous&27.586&25.806&26.667&3.12\%\\
Temporal.Synchronous&0.000&0.000&0.000&1.92\%\\
    \hline
\end{tabular}
\caption{Precision, Recall and $F_1$ scores of different labels predicted by the basic model using main evaluation metric and their proportions in test data }
\label{table:b}
\end{small}
\end{table*}


The classifier obtains relatively higher scores on some types of labels. The first type is senses with larger sample size in the corpus, suggesting the imbalanced classification problem. Two senses occur frequently in the corpus (\textsc{Contingency.Cause} and \textsc{Expansion.Conjunction}) are recognized with high Recall, but low Precision. This could indicate a strong signal, but one that is likely to be ambiguous. Other less frequent labels are constantly misclassified into these frequent labels. For example, the amount of \textsc{Expansion.Manner} samples is largely reduced by our method dealing with multi-label instances, and the classifier fails to recognize the minority class. Another type of senses achieving high scores are those occurring predominantly in intra-sentential relations (\textsc{Contingency.Purpose} and \textsc{Contingency.Condition}) or in inter-sentential relations (\textsc{Expansion.Instantiation} and \textsc{Expansion.Level-of-detail}). The model recognize these senses with high Precision, but different levels of Recall, which could be due to a difference in the strength of evidence signalling the relation. Additionally, \textsc{Temporal.Asynchronous} sense that associates with much higher proportion in linked relations than stand-alone ones obtain similar Recall and Precision scores.



\end{document}